\documentclass[twocolumn,10pt]{asme2e}

\newcommand{\be}{\begin{equation}}
\newcommand{\ee}{\end{equation}}
\newcommand{\beq}{\begin{equation}}
\newcommand{\eeq}{\end{equation}}
\newcommand{\bed}{\begin{displaymath}}
\newcommand{\eed}{\end{displaymath}}
\newcommand{\beqa}{\begin{eqnarray}}
\newcommand{\eeqa}{\end{eqnarray}}
\newcommand{\beqann}{\begin{eqnarray*}}
\newcommand{\eeqann}{\end{eqnarray*}}
\newcommand{\bseq}{\begin{subequation}}
\newcommand{\eseq}{\end{subequation}}

\newcommand{\ba}{\begin{array}}
\newcommand{\ea}{\end{array}}

\usepackage[dvips]{graphicx}
\usepackage{psfrag}
\usepackage{epsfig}
\usepackage{array}
\usepackage{amssymb}
\usepackage{subeqn}
\usepackage{setspace}
\usepackage{subfigure}

\def\confshortname{DETC'2007} 
\def\conffullname{2007 ASME Design Engineering
Technical Conferences} 
\def\confdate{4-7} 
\def\confmonth{September}
\def\confyear{2007} 
\def\confcity{Las Vegas}
\def\confstate{Nevada}
\def\confcountry{USA}

\def\papernum{DETC2007/XXX}

\title{THE MULTIOBJECTIVE OPTIMIZATION OF A PRISMATIC DRIVE}
\author{\'{E}milie Bouyer, \, St\'{e}phane Caro, \, Damien Chablat
    \affiliation{
    Institut de Recherche en Communications \\
    et Cybern\'etique de Nantes\\
    UMR CNRS n$^\circ$ 6597\\
    1 rue de la No\"e, 44321 \\
    Nantes, France\\
    Email: \{stephane.caro, damien.chablat\}@irccyn.ec-nantes.fr
    }
}

\author{Jorge Angeles
    \affiliation{Department of Mechanical Engineering \& \\
    Centre for Intelligent Machines\\
    McGill University\\
    817 Sherbrooke West, H3A 2K6, \\
    Montreal, QC, Canada \\
    Email: angeles@cim.mcgill.ca
    }
}

\begin{document}

\maketitle

\begin{abstract}
\emph{The multiobjective optimization of Slide-o-Cam is reported in
this paper. Slide-o-Cam is a cam mechanism with multiple rollers
mounted on a common translating follower. This transmission provides
pure-rolling motion, thereby reducing the friction of
rack-and-pinions and linear drives. A Pareto frontier is obtained by
means of multiobjective optimization. This optimization is based on
three objective functions: (i) the pressure angle, which is a
suitable performance index for the transmission because it
determines the amount of force transmitted to the load vs.\ that
transmitted to the machine frame; (ii) the Hertz pressure used to
evaluate the stresses produced on the contact surface between cam
and roller; and (iii) the size of the mechanism,\ characterized by
the number of cams and their width.}
\end{abstract}

\begin{nomenclature}
\entry{$p$:}{pitch of the transmission;}

\entry{$e$:}{distance between the axis of the cam and the line of
centers of the rollers;}

\entry{$r$:}{radius of the roller;}

\entry{$d_{cs}$:}{diameter of the camshaft ($d_{cs}=2(e-r)$);}

\entry{$L$:}{the width of the contact between the cams and the
rollers;}

\entry{$\psi$:}{\emph{input} of the mechanism, i.e., the angle of
rotation of the cam;}

\entry{$s$:}{\emph{output} of the mechanism, i.e., the displacement
of the follower;}

\entry{$\mu$:}{pressure angle;}

\entry{\textbf{f}:}{force transmitted from the cam to the roller;}

\entry{$\kappa_{c}$ and $\kappa_{p}$:}{curvature of the cam profile
and the pitch curve, respectively;}

\entry{$\rho_{c}$ and $\rho_{p}$:}{radii of curvature of the cam
profile and the pitch curve, respectively;}

\entry{$m$:}{number of cams mounted on the camshaft;}

\entry{$n$:}{number of lobes per cam;}

\entry{$P$:}{Hertz pressure;}

\entry{$S_M$:}{size of the mechanism.}

\end{nomenclature}

\section{INTRODUCTION}
In robotic and mechatronic applications, whereby motion is
controlled using a piece of software, the conversion from rotational
to translational motion is usually realized by means of {\em
ball-screws} or {\em linear actuators}. While both are gaining
popularity, they present some drawbacks. On the one hand,
ball-screws comprise a high number of moving parts, their
performance depending on the number of balls rolling in the shaft
groove. Moreover, they have a low load-carrying capacity due to the
punctual contact between balls and groove. On the other hand, linear
bearings are composed of roller-bearings to figure out the previous
issue, but these devices rely on a form of direct-drive motor, which
makes them expensive to produce and maintain.

\begin{figure}[htb]
 \begin{center}
   \psfrag{Roller}{Roller}
   \psfrag{Follower}{Follower}
   \psfrag{Conjugate cams}{Conjugate cams}
   \epsfig{file=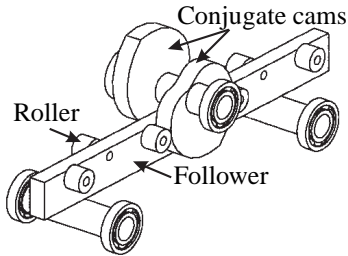,scale = 0.4}
   \caption{Layout of Slide-o-Cam}
   \label{fig001}
 \end{center}
\end{figure}

A novel transmission, called {\it Slide-o-Cam},\ is depicted in
Fig.~\ref{fig001} as introduced in \cite{Gonzalez-Palacios:2000} to
transform a rotational motion into a translational one. Slide-o-Cam
is composed of four main elements: ($i$) the frame; ($ii$) the cam;
($iii$) the follower; and ($iv$) the rollers. The input axis on
which the cams are mounted, named \emph{camshaft}, is driven at a
constant angular velocity by means of an actuator under
computer-control. Power is transmitted to the output, the
translating follower, which is the roller-carrying slider, by means
of pure-rolling contact between the cams and the rollers. The roller
comprises two components, the pin and the bearing. The bearing is
mounted to one end of the pin, while the other end is press-fit into
the roller-carrying slider. Consequently, the contact between the
cams and rollers occurs at the outer surface of the bearing. The
mechanism uses two conjugate cam-follower pairs, which alternately
take over the motion transmission to ensure a positive action; the
rollers are thus driven by the cams throughout a complete cycle.
Therefore, the main advantages of cam-follower mechanisms with
respect to the other transmissions, which transform rotation into
translation are: ($i$)~lower friction; ($ii$)~higher stiffness;
($iii$)~low backlash; and ($iv$)~reduction of wear.
The multiobjective optimization of Slide-o-Cam is reported in this
paper. This optimization is based on three criteria: ($i$) the
pressure angle,\ a suitable performance index for the transmission
because it determines the amount of force transmitted to the load
vs.\ that transmitted to the machine frame; ($ii$) the Hertz
pressure,\ a measure of the stresses produced in the contact surface
between the cams and the rollers; and ($iii$) the size of the
mechanism,\ characterized by the number of cams and their width.

\section{SYNTHESIS OF PLANAR CAM MECHANISMS}
\begin{figure}[htb]
 \begin{center}
   \psfrag{f}{$\bf f$}
   \psfrag{p}{$p$}    \psfrag{e}{$e$}    \psfrag{p}{$p$}    \psfrag{s}{$s$}
   \psfrag{d}{$d$}    \psfrag{x}{$x$}   \psfrag{y}{$y$}    \psfrag{P}{$P$}
   \psfrag{C}{$C$}    \psfrag{mu}{$\mu$}
   \psfrag{u}{$u$}       \psfrag{v}{$v$}
   \psfrag{delta}{$\delta$}
   \psfrag{b2}{$b_1$}   \psfrag{b3}{$b_2$}
   \psfrag{a4}{$r$}   \psfrag{psi}{$\psi$}
   \psfrag{theta}{$\theta$}
   \psfrag{O1}{$O_1$}   \psfrag{O2}{$O_2$}
   \centerline{\epsfig{file = 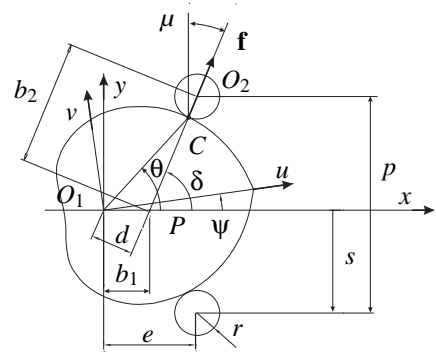,scale = 0.3}}
  \caption{Parameterization of \goodbreak Slide-o-Cam}
  \label{fig002}
 \end{center}
 \end{figure}

Let the $x$-$y$ frame be fixed to the machine frame and the $u$-$v$
frame be attached to the cam, as depicted in Fig.~\ref{fig002}.
$O_{1}$ is the origin of both frames, $O_{2}$ is the center of the
roller, and $C$ is the contact point between cam and roller.
 \begin{figure}[htb]
 \begin{center}
   \psfrag{O1}{$O_1$}
   \psfrag{p}{$p$}    \psfrag{x}{$x$}    \psfrag{y}{$y$}
   \psfrag{u}{$u$}    \psfrag{v}{$v$}    \psfrag{x}{$x$}
   \psfrag{s(0)}{$s(0)$}
   \centerline{\epsfig{file = 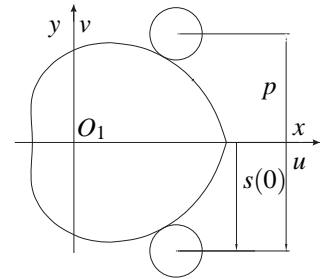,scale = 0.3}}
   \caption{Home configuration of the mechanism} \label{fig003}
 \end{center}
 \end{figure} The geometric parameters are illustrated in the same figure. The
notation used in this figure is based on the general notation
introduced in \cite{Waldron:1999,Carra:2004}, namely, ($i$)~the
pitch $p$, {\it i.e.}, the distance between the center of two
rollers on the same side of the follower; ($ii$)~the distance $e$
between the axis of the cam and the line of centers of the rollers;
($iii$)~the radius $r$ of the roller-bearing, {\it i.e.}, the radius
of the roller; ($iv$)~the angle of rotation $\psi$ of the cam, the
input of the mechanism; ($v$)~the position $s$ of the center of the
roller, {\it i.e}, the displacement of the follower, which is the
output of the mechanism; ($vi$)~the pressure angle $\mu$; and
($vii$)~the force {\bf f} transmitted from the cam to the roller.

The above parameters as well as the surface of contact on the cam
are determined by the geometric relations derived from the
Aronhold-Kennedy Theorem~\cite{Waldron:1999}. As a matter of fact,
when the cam makes a complete turn, i.e., $\Delta\psi=2\pi$, the
displacement of the roller is equal to the pitch, i.e., $\Delta
s=p$. Furthermore, if we consider that Fig.~\ref{fig003} illustrates
the home configuration of the roller, the latter is below the
$x$-axis when $\psi=0$. Therefore, $s(0)=-p/2$ and the input-output
function $s$ is defined as:
\begin{equation}
  s(\psi)=\frac{p}{2\pi}\psi-\frac{p}{2} \label{eq01}
 \end{equation}
The cam profile is determined by the displacement of the contact
point $C$ around the cam. The Cartesian coordinates of $C$ in the
$u$-$v$ frame take the form~\cite{Gonzalez-Palacios:1993}
\begin{subequations}
  \begin{eqnarray}
    u_c(\psi)&=&b_{1} \cos \psi+(b_2-r)\cos(\delta-\psi)
    \label{eq:ucphi}\\
    v_{c}(\psi)&=&-b_{1} \sin \psi + (b_{2}-r)\sin(\delta-\psi)
    \label{eq:vcphi}
  \end{eqnarray}
\end{subequations}
The expressions of coefficients $b_{2}$, $b_{3}$ and $\delta$,\ as
obtained in~\cite{Lee:2001,Chablat:2006},\ are:
\begin{subequations}
  \begin{eqnarray}
    b_{1}&=&\frac{p}{2\pi}
    \label{eq:b2}\\
    b_{2}&=&\frac{p}{2\pi}\sqrt{(2\pi \eta -1)^{2}+(\psi-\pi)^{2}}
    \label{eq:b3}\\
    \delta&=&\arctan\left(\frac{\psi-\pi}{2\pi \eta -1} \right)
    \label{eq:delta}
  \end{eqnarray}
\end{subequations} where $\eta=e/p$,\ a nondimensional design parameter.

From Eq.(\ref{eq:delta}), we can notice that $\eta$ cannot be equal
to $1/(2\pi)$. \vspace{-0.7cm}
\begin{figure}[!ht]
 \begin{center}
     \psfrag{O1}{$O_1$}
     \psfrag{x}{$x$}
     \psfrag{y}{$y$}
     \psfrag{u}{$u$}
     \psfrag{v}{$v$}
     \psfrag{C}{$C$}
     \psfrag{Psi}{$\psi$}
     \subfigure[]{\epsfig{file=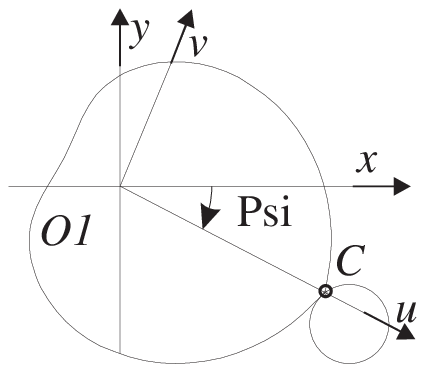,scale = 0.6}}
     \psfrag{x, u}{$x, u$}
     \psfrag{y, v}{$y, v$}
     \psfrag{delta}{$\Delta$}
     \psfrag{-u}{-$u$}
     \psfrag{-v}{-$v$}
     \subfigure[]{\epsfig{file=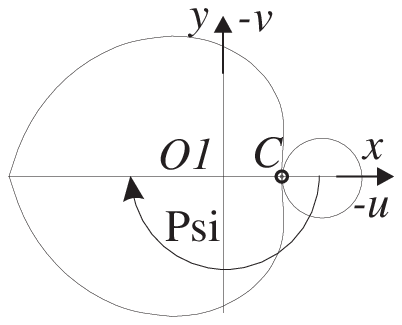,scale = 0.6}}
     \psfrag{u}{$u$}
     \psfrag{v}{$v$}
     \subfigure[]{~~~\epsfig{file=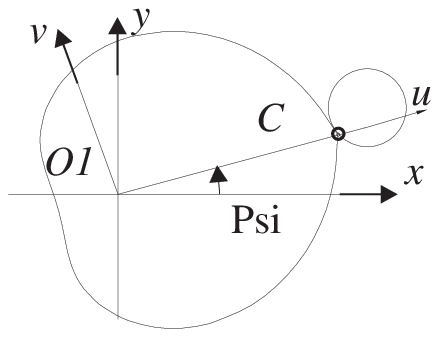,scale = 0.6}}
 \end{center}
 \vspace{-0.7cm}
 \caption{Orientations of the cam found when $v_c=0$: (a) $\psi=\Delta$;
  (b) $\psi=\pi$; and (c) $\psi=2\pi-\Delta$}
 \label{fig004}
 \end{figure} \vspace{-0.7cm} Moreover, an {\it extended angle} $\Delta$ was
introduced in~\cite{Lee:2001} to obtain a closed cam profile.
$\Delta$ is defined as a root of Eq.(\ref{eq:vcphi}). As far as
Slide-o-Cam is concerned, $\Delta$ is negative, as shown in
Fig.~\ref{fig004}. Consequently, to close the cam profile,\ $\psi$
must vary within $\Delta \leq \psi \leq 2\pi-\Delta$.

\subsection{Pitch-Curve Determination}
The pitch curve is the trajectory of $O_{2}$, the center of the
roller, distinct from the trajectory of the contact point $C$, which
produces the cam profile. The Cartesian coordinates $(e,s)$ of
point~$O_{2}$ in the $x$-$y$ frame are depicted in
Fig.~\ref{fig002}. Hence, the Cartesian coordinates of the
pitch-curve in the $u$-$v$ frame are
 \begin{subequations}
 \begin{eqnarray}
    u_{p}(\psi)&=& e \cos \psi + s(\psi)\sin \psi
    \label{eq:up}\\
    v_{p}(\psi)&=& -e \sin \psi + s(\psi)\cos \psi \label{eq:vp}
  \end{eqnarray}
 \end{subequations}
\subsection{Curvature of the Cam Profile}
The curvature $\kappa_{p}$ of the pitch curve is given in
\cite{Chablat:2006} as
 \begin{equation}
 \label{eq6}
 \kappa_{p}=\frac{2\pi}{p} \frac{[(\psi-\pi)^{2}+2(2\pi
\eta-1)(\pi \eta-1)]}{[(\psi-\pi)^{2}+(2\pi \eta-1)^{2}]^{3/2}}
 \end{equation}
provided that the denominator does not vanish at any value of $\psi$
within $\Delta \leq \psi \leq 2\pi-\Delta$, i.e.,
$\eta\neq1/(2\pi)$.

Let $\rho_{c}$ and $\rho_{p}$ be the radii of curvature of the cam
profile and the pitch curve, respectively, and $\kappa_{c}$ the
curvature of the cam profile. Since the curvature is the reciprocal
of the radius of curvature, we have $\rho_{c} = 1/\kappa_{c}$ and
$\rho_{p} = 1/\kappa_{p}$. Furthermore, due to the definition of the
pitch curve, it is apparent that
 \begin{equation} \label{eq8}
    \rho_{p} = \rho_{c} + r
 \end{equation}
From Eq.~(\ref{eq8}), the curvature of the cam profile can be
written as
 \begin{equation} \label{eq:kappaC}
 \kappa_{c}=\frac{\kappa_{p}}{1-r \kappa_{p}}
 \end{equation}
In \cite{Chablat:2005}, the authors claimed that the cam profile has
to be fully convex for machining accuracy. Such a profile can be
obtained if and only if $\eta>1/\pi$. In order to increase the range
of design parameters, we include non-convex cams within the scope of
this paper. Nevertheless, the sign of the local radius $\rho_c$ has
to remain positive as long as the cam pushes the roller. In this
vein, the cam is convex when $\eta \in ]1/(2\pi), \, 1/\pi]$ and
$\psi \in ]\Delta, \, \pi]$~\cite{Chablat:2007}.
\begin{figure}[!h]
  \begin{center}
 \psfrag{psi}{$\psi$}
 \psfrag{kappa}{$1/\kappa_c$}
 \psfrag{x}{$x$}
 \psfrag{y}{$y$}
 \psfrag{15}{{\tiny 15}}
 \psfrag{10}{{\tiny 10}}
 \psfrag{5}{{\tiny 5}}
 \psfrag{0}{{\tiny 0}}
 \psfrag{-5}{{\tiny -5}}
 \psfrag{-1}{{\tiny -1}}
 \psfrag{1}{{\tiny 1}}
 \psfrag{2}{{\tiny 2}}
 \psfrag{3}{{\tiny 3}}
  \includegraphics[width=8cm]{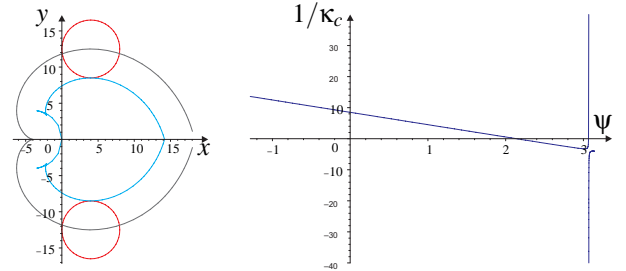}
  \caption{Cam profile and local curvature of the cam}
  \label{fig005}
  \end{center}
\end{figure} Moreover, according to~\cite{Chablat:2005}, $\rho_c$ is a minimum when
\begin{equation}\label{eq:psimin}
    \psi = \psi_{min}= {\frac {\pi -\sqrt {4\,{n}^{2}\pi \,h-{n}^{2}-4\,{n}^{2}{\pi }^{2}{h}^{2}}}{n}}
\end{equation} where $n$ is the number of lobes per cam. Therefore, the cam profile is not feasible when
$\rho_c(\psi_{min})<0$. If this inequality becomes an equality, the
roller will block the cam,\ as depicted in Fig.~\ref{fig005}.

\section{MULTIOBJECTIVE OPTIMIZATION PROBLEM}
We introduce in this section the multiobjective optimization of
Slide-o-Cam. Indeed, such an optimization is needed to properly
dimension the mechanism. First, the objective functions are defined.
Then, a sensitivity analysis of the mechanism is reported in order
to choose shrewdly the design variables of the optimization problem.
Finally, the results of the latter are illustrated by means of a
Pareto frontier as the objective functions are antagonistic.
\subsection{The Objective Functions}
The optimization of the mechanism is based on three objective
functions: $(i)$~the maximum pressure angle $\mu_{max}$; $(ii)$~the
maximum Hertz pressure $P_{max}$ related to the contact between the
cams and the rollers; and $(iii)$~the size of the mechanism~$S_M$.
As a matter of fact, we want to simultaneously minimize these three
functions.
\subsubsection{The Pressure Angle}
\label{sec:mu} The pressure angle $\mu$ of a cam-roller-follower
mechanism is defined as the angle between the normal to the contact
point $C$ between the cam and the roller and the velocity of $C$ as
a point of the follower~\cite{Angeles:1991}. As illustrated in
Fig.~\ref{fig002}, $\mu$ is a significant parameter in cam design.
In fact, the smaller $\mu$\footnote{$\mu$ is a real number and can
be either positive or negative. However, within the scope of this
paper, $\mu$ remains positive. Therefore, $\mu=|\,\mu\,|$,
$|\,\cdot\,|$ denoting the absolute value.}, the better the
transmission. The expression for $\mu$ is given
in~\cite{Angeles:1991}; in terms of $\eta$, we have
\begin{equation}\label{eq:mu}
    \tan\mu=\frac{n-2n\pi\eta}{n\psi-\pi}
\end{equation}

\begin{figure}[!h]
\centering
  \psfrag{toto}[l][l][0.75]{active part}
  \psfrag{mumax}[l][l][0.75]{$\mu=\mu_{max}$}
  \psfrag{phmax}[l][l][0.75]{$P=P_{max}$}
  \psfrag{a}{(a)}
  \psfrag{b}{(b)}
  \psfrag{x}{$x$}
  \psfrag{y}{$y$}
  \includegraphics[width=7.5cm]{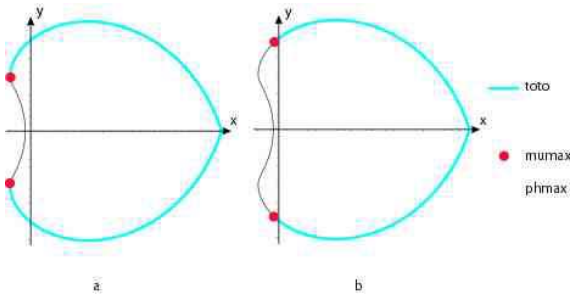}\\
  \caption{Active parts of: (a) a two- and (b) three-conjugate cam mechanisms}\label{fig:activeparts}
\end{figure}

Figure~\ref{fig:activeparts} illustrates the active parts of a two-
and a three-conjugate-cam mechanisms. It turns out that the pressure
angle is a maximum at the ends of the active parts for the two
mechanisms. In this paper, $\mu_{max}$ denotes the maximum pressure
angle along the active part of the cam profile; it is an objective
function in this optimization problem.

\subsubsection{The Hertz Pressure}
\label{sec:PHz} When two bodies with curved surfaces, for example, a
cam and a roller, are pressed together, contact takes place not
along a line but along a surface, due to the inherent material
compliance. Moreover, the stresses developed in the two bodies are
three-dimensional. Those contact stresses may generate failures as
cracks, pits, or flaking in the surface material.
\begin{figure}[!h]
  \begin{center}
\psfrag{x}{$x$} \psfrag{y}{$y$} \psfrag{B}{$B$} \psfrag{O1}{$O_1$}
\psfrag{d}{$\rho_{c}$} \psfrag{L}{$L$} \psfrag{r}{$r$}
  \includegraphics[width=5.5cm]{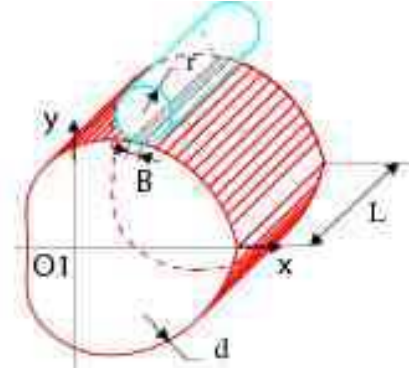}
  \caption{The width $B$ of the contact between a cam and a roller}
  \label{fig:bandwidthB}
  \end{center}
\end{figure} To quantify these stresses, Heinrich Rudolf
Hertz~(1857--1894) proposed some formulas to evaluate the width of
the band of contact between two cylinders and the maximum pressure
of contact, called \emph{Hertz pressure}. In Slide-o-Cam, the
rollers and the cams are the bodies in contact. Unlike the roller,
the cam is not a cylinder, but can be approximated by a cylinder
with radius identical to the radius of curvature of the cam at the
contact point. The width $B$ of the band of contact is illustrated
in Fig.\ref{fig:bandwidthB}, and given by Hertz as
\begin{subequations}
 \begin{eqnarray}
    B&=&\sqrt{\frac{16F(K_1+K_2)R_{equ}}{L}}
    \label{e:B-hertz-formula}\\
    R_{equ}&=&\frac{r\,\rho_c}{r+\rho_c} \label{eq:Requ}
  \end{eqnarray}
 \end{subequations} $F$ being the magnitude of the axial load \textbf{f} while $R_{equ}$ is the equivalent radius of
 contact, $L$ the width of the contact between the cams and the
 rollers, and $K_1$ and $K_2$ the coefficients that characterize the materials of the cams and the rollers,
 respectively, i.e.,
\begin{equation}
    K_1=\frac{1-\nu_1^2}{\pi E_1}, \quad K_2=\frac{1-\nu_2^2}{\pi E_2}
\end{equation} where $\nu_1$ and $\nu_2$ are the Poisson ratios of the materials of the cam
and the roller, respectively, while $E_1$, $E_2$ their corresponding
Young moduli. Accordingly, the Hertz pressure $P$ of the contact
between the cams and the rollers takes the form:
\begin{equation}
   \label{e:pmax-hertz-formula}
   P=\frac{4F}{L \pi B}
\end{equation} Let us notice that $P$ depends on $\psi$, as $F$ is a
function of this variable and $B$ is a function of $F$.

Let us assume that $F$ is constant. As $L$ and $r$ are constant and
$R_{equ}$ is monotonic with respect to (w.r.t) $\rho_c$ as long as
$\rho_c>-r$, from Eq.(\ref{e:B-hertz-formula}), the lower $\rho_c$,
the lower $B$. From Eq.(\ref{e:pmax-hertz-formula}), the lower $B$,
the higher $P$. According to \cite{Chablat:2007}, $\rho_c$ is a
minimum when $\psi=~\pi/n-\Delta$ for a two-conjugate cam mechanism.
Therefore, $P$ is a maximum when $\psi=~\pi/n-\Delta$ for such a
mechanism.

Figure~\ref{fig:activeparts} illustrates the active parts of a two-
and a three-conjugate-cam mechanisms. It turns out that the Hertz
pressure is a maximum at the ends of the active parts for the two
mechanisms as $\rho_c$ is a minimum at those ends. In this paper,
$P_{max}$ denotes the maximum Hertz pressure along the active part
of the cam profile;\ it is an objective function in this
optimization problem.

The maximum Hertz pressures allowed for some materials are obtained
from \cite{matweb:2007} and recorded in
Table~\ref{t:allowable-pressure}. The second column gives the
allowable pressure $P_{stat}$ for a static load. As a matter of
fact, it is recommended not to apply more than $40\%$ of $P_{stat}$
in order to secure an infinite fatigue life. The corresponding
values $P_{max}$ are given in the third column of
Table~\ref{t:allowable-pressure}.
\begin{table}[!htbp]
\begin{center}
\caption{Allowable pressures}\label{t:allowable-pressure}
{\renewcommand{\arraystretch}{1}
\begin{tabular}{ccc}
  \hline \hline
  Material & $P_{stat}$~[MPa] & $P_{max}$~[MPa]  \\
  \hline
    Stainless steel   & 650 & 260 \\
    Improved steel &  1600 to 2000 & 640 to 800 \\
    Grey cast iron  & 400 to 700 & 160 to 280 \\
    Aluminum & 62.5 & 25 to 150 \\
    Polyamide & 25 & 10 \\
  \hline \hline
\end{tabular}}
\end{center}
\end{table}
Obviously, the maximum allowable pressure depends also on the shape
of the different parts in contact. A thick part will be stiffer than
a thin one. Nevertheless, we only take into account the material of
the cams and rollers for the determination of the allowable
pressures within the scope of this research work. Finally, let us
notice that only improved steel is appropriate for a Slide-o-Cam
transmission in case of high Hertz-pressure values.
\subsubsection{Size}
The size of the mechanisms $S_M$ is defined as
\begin{equation}\label{eq:SM}
    S_M = m \, L
\end{equation} where $m$ is the number of cams. From \cite{Chablat:2005}, a Slide-o-Cam with
only one cam, i.e., $m=1$, is not feasible. Besides, the smaller
$S_M$, the less bulky the mechanism.

\subsection{The Design Variables}
The design variables of the optimization problem are: ($i$)~the
diameter $d_{cs}$ of the camshaft ($d_{cs}=e-r$); ($ii$)~the radius
$r$ of the rollers; ($iii$)~the width $L$ of the contact between cam
and roller; and ($iv$)~the number of cams~$m$.
\subsection{Sensitivity Analysis}
We conduct here the analysis of the sensitivity of the performance
of Slide-o-Cam to the variations in its design parameters. Such an
analysis is needed to both determine the tolerance of the design
variables and obtain a robust design.
\subsubsection{Sensitivity of the Pressure Angle}
\begin{figure}[!h]
\centering
 \psfrag{mu}[c][c][0.75]{$\mu$ (degree)}
 \psfrag{psi}[c][c][0.75]{$\psi$(rad)}
 \psfrag{12}[c][c][0.5]{12} \psfrag{10}[c][c][0.5]{10}   \psfrag{8}[c][c][0.5]{8} \psfrag{7}[c][c][0.5]{7}
 \psfrag{6}[c][c][0.5]{6}   \psfrag{5}[c][c][0.5]{5}     \psfrag{4}[c][c][0.5]{4} \psfrag{3}[c][c][0.5]{3}
 \psfrag{2}[c][c][0.5]{2}   \psfrag{0}[c][c][0.5]{0}
 \psfrag{-2}[c][c][0.5]{-2} \psfrag{-4}[c][c][0.5]{-4}   \psfrag{-6}[c][c][0.5]{-6}
 \psfrag{-20}[c][c][0.5]{-20}  \psfrag{-40}[c][c][0.5]{-40}  \psfrag{-60}[c][c][0.5]{-60}  \psfrag{-80}[c][c][0.5]{-80}
 \psfrag{20}[c][c][0.5]{20}    \psfrag{40}[c][c][0.5]{40}    \psfrag{60}[c][c][0.5]{60}    \psfrag{80}[c][c][0.5]{80}
 \psfrag{t_c_p}[c][c][0.75]{the came pushes}
 \psfrag{t_t_l}[c][c][0.75]{to the left}
 \psfrag{t_t_r}[c][c][0.75]{to the right}
 \psfrag{(A)}{(a)}
 \psfrag{(B)}{(b)}
  \includegraphics[width=8cm]{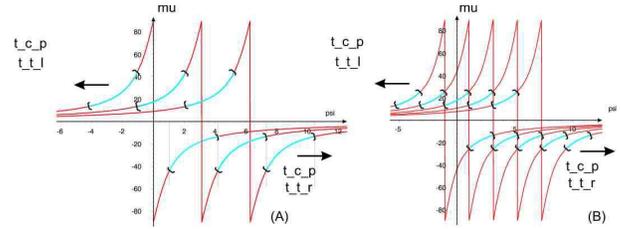}\\
  \caption{Pressure-angle distribution for (a) two conjugate-cam and (b) three conjugate-cam mechanisms with one lobe, $p=50$, $r=10$ and $e=9$}
  \label{fig:comptwothreeconjcams}
\end{figure}
Figure~\ref{fig:comptwothreeconjcams} illustrates the pressure-angle
distribution for two conjugate- and three conjugate-cams with one
lobe. We can notice that the pressure angle decreases with the
number of cams. Consequently, we can use conjugate cams, namely,
several cams mounted on the camshaft, to reduce the pressure angle.
Below is a list of the effects of some design parameters on the
pressure angle:
\begin{enumerate}
  \item The lower $\eta$, the lower the pressure angle, with $\eta \geq
  1/\pi$;
  \item the lower $r$, the lower the pressure angle;
  \item the lower $n$, the lower the pressure angle,\,\cite{Chablat:2005};
  \item the higher $m$, the lower the pressure angle.
\end{enumerate} where $m$ is the number of cam(s) mounted on the
camshaft.

As the pressure angle increases with the number of lobes, we
consider only mono-lobe cams, i.e. $n=1$.
\subsubsection{Sensitivity of the Hertz Pressure} $P_{max}$ depends on the geometry of the cam, the number of
conjugate cams, the material of the parts in contact and the load
applied. Therefore, we have different ways to minimize the Hertz
pressure, namely,
\begin{enumerate}
  \item The higher $m$, the lower $P_{max}$;
  \item the lower the axial load, the lower $P_{max}$;
  \item the more compliant the material, the lower $P_{max}$;
  \item the higher $L$, the lower $P_{max}$.
\end{enumerate}
In order to analyze the sensitivity of $P$ to $r$, $\eta$, $p$ and
$L$, we use a first derivative model of $P$ w.r.t.\ the
corresponding parameters, i.e.,
\begin{equation}\label{e:deltaPHz}
    \delta P \approx \textbf{c} \, \delta \textbf{q}
\end{equation}
with
\begin{equation}\label{e:c}
    \textbf{c}~=~\left[
\begin{array}{c}
  \displaystyle{{\partial P}/{\partial r}} \\
  \displaystyle{{\partial P}/{\partial \eta}} \\
  \displaystyle{{\partial P}/{\partial p}} \\
  \displaystyle{{\partial P}/{\partial L}}
\end{array} \right] \, , \, \delta \textbf{q}~=~\left[
\begin{array}{c}
  \delta r \\
  \delta \eta \\
  \delta p \\
  \delta L
\end{array}
\right]
\end{equation}
If the values of the parameters are known, we will be able to
evaluate \textbf{c}. Let us assume that $r=4$~mm, $\eta=0.18$,
$p=50$~mm and $L=10$~mm.
\begin{figure}[!htbp]
  \centering
  \psfrag{dqi}[c][c]{$\displaystyle{\frac{\partial P}{\partial q_i}\times q_{i0}}$}
  \psfrag{psi}[c][c]{$\psi$}
  \psfrag{a4}[l][]{w.r.t $r$}
  \psfrag{h}[l][]{w.r.t $\eta$}
  \psfrag{p}[l][]{w.r.t $p$}
  \psfrag{L}[l][]{w.r.t $L$}
  \psfrag{Cm}[l][]{w.r.t $C_m$}
  \psfrag{0}[r][c]{$0$}
  \psfrag{5.0}[c][c]{$5.0$}
  \psfrag{6.0}[c][c]{$6.0$}
  \psfrag{7.0}[c][c]{$7.0$}
  \psfrag{200}[r][c]{$200$}
  \psfrag{100}[r][c]{$100$}
  \psfrag{-100}[r][c]{$-100$}
  \psfrag{-200}[r][c]{$-200$}
  \psfrag{-300}[r][c]{$-300$}
  \psfrag{a}[c][c]{$\psi=\pi/n-\Delta$}
  \psfrag{b}[c][c]{$\psi=2\pi/n-\Delta$}
  \includegraphics[width=8cm]{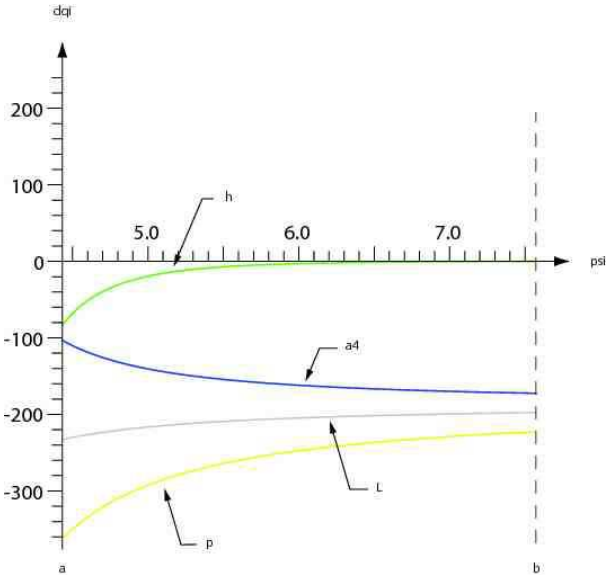}\\
  \caption{Influence of the variations in $r$, $\eta$, $p$ and $L$ on $P$}\label{f:sensitivity-analysis}
\end{figure}
The partial derivatives have to be normalized to be compared. In
this vein, we divide each of them by its nominal value. Now, we can
plot each partial derivative with respect to the angle of rotation
of the cam $\psi$, as illustrated in
Fig.~\ref{f:sensitivity-analysis}.

The most influential variables are those with the highest absolute
value of their corresponding partial derivative for a given value of
$\psi$. As the maximum value of the Hertz pressure is obtained for
$\psi=\pi/n-\Delta$ and $\Delta=-1.2943$~rad, the partial
derivatives can be evaluated for $\psi=\pi/n-\Delta$. The
sensitivity of $P_{max}$ to $\delta \textbf{q}$ is recorded in
Table~\ref{t:variable-Pmax}.
\begin{table}[!htbp]
  \centering
  \caption{Influence of the variations in $r$, $\eta$, $p$ and $L$ on $P_{max}$}
  \begin{tabular}{ccccc}
              \hline\hline
              $q_i$ & $r$ & $\eta$ & $p$ & $L$  \\
              \hline
              $q_{init}$ & 4~mm & 0.18 & 50~mm  & 10~mm \\
              $\displaystyle{\left|\frac{\partial P_{max}}{\partial q_i}(q_{init})\right|}$
              & 103.32 & 83.25 & 362.03 & 232.67  \\
              Order of importance & 3 & 4 & 1 & 2 \\
              \hline\hline
            \end{tabular}
  \label{t:variable-Pmax}
\end{table}
The plots in Fig.~\ref{f:sensitivity-analysis} show the sensitivity
of the Hertz pressure w.r.t the different parameters for different
values of $\psi$ and for the active part of the cam profile.
However, it is more relevant to calculate the rms value of each
partial derivative, as recorded in Table~\ref{t:variable-Ph}.
\begin{table}[!htbp]
  \centering
  \caption{Global influence of the variations in $r$, $\eta$, $p$ and $L$ on $P$}
  \begin{tabular}{ccccc}
              \hline\hline
              $q_i$ & $r$ & $\eta$ & $p$ & $L$  \\
              \hline
              $q_{init}$ & 4  & 0.18 & 50  & 10   \\
              $\sqrt{\displaystyle\frac{n}{\pi}\int_{\psi}(\displaystyle{\frac{\partial P}{\partial q_i}})^2{\rm d}\psi}$
              & 156.59 & 20.21 & 261.85 & 207.79  \\
              Order of importance & 3 & 4 & 1 & 2 \\
              \hline\hline
            \end{tabular}
  \label{t:variable-Ph}
\end{table}
As a matter of fact, Table~\ref{t:variable-Pmax} and
Table~\ref{t:variable-Ph} provide the same results in terms of order
of importance of the variations in $r$, $\eta$, $p$ and $L$.
Finally, in order to minimize the variations in the Hertz pressure,
we had better minimize the variations in $p$, $L$ $r$ and $\eta$ in
descending order.

\subsubsection{Sensitivity of the Size of the Mechanism} The
sensitivity analysis of $S_M$ is trivial. Indeed, from
eq.~(\ref{eq:SM}), the higher $m$, the higher $S_M$. Likewise, the
higher $L$, the higher $S_M$.

\subsection{Problem Formulation}
\label{sec:pbformulation} A motivation of this research work is to
implement a Slide-o-Cam transmission in the Orthoglide, a low-power
machine tool introduced in \cite{Chablat:2003}. To that end, the
transmission has to transmit a torque $C_t$ of 1.2~Nm with a pitch
of 20~mm. In case of high-speed operations, i.e., when the velocity
of the cams is higher than 50~rpm, the pressure-angle is recommended
to be smaller than $30^\circ$. Table~\ref{t:allowable-pressure}
shows that the maximum value of the Hertz pressure has to be smaller
than 800~MPa as the cams and the rollers are made up of steel.
Moreover $S_M$ is supposed to be smaller than 90~mm with a view to
limiting the size of the mechanism. Besides, the Slide-o-Cam
transmissions under study are composed of two- or three- conjugate
cams as a Slide-o-Cam with only one cam is not feasible and such a
mechanism with more than three conjugate cams would be too bulky,
i.e., $m=\{2,3\}$. Consequently, the optimization problem can be
formulated
\[ \left.
\begin{array}{l}
\quad \displaystyle{\min_{{\bf x}}\left(\mu_{max}, \, P_{max}, \, S_M \right)}\\
s.t.\\
\quad \mu_{max} \leq 30^\circ\\
\quad P_{max} \leq 800~\textrm{MPa}\\
\quad S_M \leq 90~\textrm{mm}\\
\quad {\bf x}_{l} \leq {\bf x} \leq {\bf x}_{u}\\
\end{array}
\right.\] where ${\bf x}=\left[d_{cs},\,r,\,L,\,m\right]^T$, while
${\bf x}_{l}$ and ${\bf x}_{u}$ denote the lower and upper bounds of
the design variables, respectively. Here, ${\bf
x}_{l}=\left[0~\textrm{mm},\,4~\textrm{mm},\,0~\textrm{mm},\,2\right]$
and ${\bf x}_{u}=\left[0~\textrm{mm},\,10.5~\textrm{mm},\,L_{max},\,
3\right]$, $L_{max}$ being equal to $S_{Max}/m$ knowing that
$S_{Max}=90~\textrm{mm}~$.

\subsection{Results}
\begin{figure}[!h]
\centering
\psfrag{s05}[B][B]{$\mu_{max}$~[deg]}%
\psfrag{s06}[rt][rt]{$S_M$~[m]}%
\psfrag{s07}[b][b]{$P_{max}$~[MPa]}%
\psfrag{toto}[c][c][0.75]{Two-conjugate cams}%
\psfrag{titi}[c][c][0.75]{Three-conjugate cams}%
%
\psfrag{x01}[t][t][0.75]{5}%
\psfrag{x02}[t][t][0.75]{10}%
\psfrag{x03}[t][t][0.75]{15}%
\psfrag{x04}[t][t][0.75]{20}%
\psfrag{x05}[t][t][0.75]{25}%
\psfrag{x06}[t][t][0.75]{30}%
\psfrag{x07}[t][t][0.75]{5}%
\psfrag{x08}[t][t][0.75]{10}%
\psfrag{x09}[t][t][0.75]{15}%
\psfrag{x10}[t][t][0.75]{20}%
\psfrag{x11}[t][t][0.75]{25}%
\psfrag{x12}[t][t][0.75]{30}%
%
\psfrag{v01}[r][r][0.75]{0}%
\psfrag{v02}[r][r][0.75]{0.01}%
\psfrag{v03}[r][r][0.75]{0.02}%
\psfrag{v04}[r][r][0.75]{0.03}%
\psfrag{v05}[r][r][0.75]{0.04}%
\psfrag{v06}[r][r][0.75]{0.05}%
\psfrag{v07}[r][r][0.75]{0.06}%
\psfrag{v08}[r][r][0.75]{0.07}%
\psfrag{v09}[r][r][0.75]{0.08}%
\psfrag{v10}[r][r][0.75]{0.09}%
\psfrag{v11}[r][r][0.75]{0.1}%
\psfrag{v12}[r][r][0.75]{0}%
\psfrag{v13}[r][r][0.75]{0.02}%
\psfrag{v14}[r][r][0.75]{0.04}%
\psfrag{v15}[r][r][0.75]{0.06}%
\psfrag{v16}[r][r][0.75]{0.08}%
\psfrag{v17}[r][r][0.75]{0.1}%
%
\psfrag{z01}[r][r][0.75]{400}%
\psfrag{z02}[r][r][0.75]{600}%
\psfrag{z03}[r][r][0.75]{800}%
\psfrag{z04}[r][r][0.75]{400}%
\psfrag{z05}[r][r][0.75]{500}%
\psfrag{z06}[r][r][0.75]{600}%
\psfrag{z07}[r][r][0.75]{700}%
\psfrag{z08}[r][r][0.75]{800}%
  \includegraphics[width=8cm]{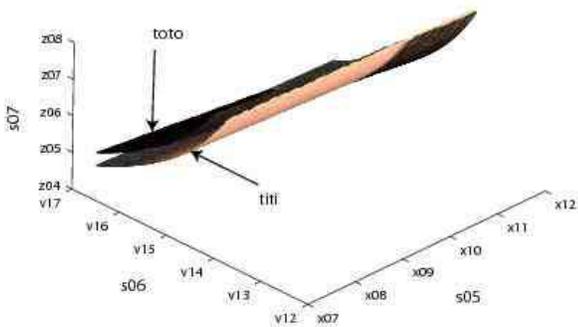}\\
  \caption{Pareto frontier of a two- and a three- conjugate cam mechanisms}
  \label{fig:3DPareto23cams}
\end{figure}

\begin{figure}[!h]
\centering
\psfrag{s05}[B][B]{$\mu_{max}$~[deg]}%
\psfrag{s06}[rt][rt]{$S_M$~[m]}%
\psfrag{s07}[b][b]{$P_{max}$~[MPa]}%
%
\psfrag{x01}[t][t][0.75]{0}%
\psfrag{x02}[t][t][0.75]{5}%
\psfrag{x03}[t][t][0.75]{10}%
\psfrag{x04}[t][t][0.75]{15}%
\psfrag{x05}[t][t][0.75]{20}%
\psfrag{x06}[t][t][0.75]{25}%
\psfrag{x07}[t][t][0.75]{30}%
\psfrag{x08}[t][t][0.75]{5}%
\psfrag{x09}[t][t][0.75]{10}%
\psfrag{x10}[t][t][0.75]{15}%
\psfrag{x11}[t][t][0.75]{20}%
\psfrag{x12}[t][t][0.75]{25}%
\psfrag{x13}[t][t][0.75]{30}%
%
\psfrag{v01}[r][r][0.75]{0}%
\psfrag{v02}[r][r][0.75]{0.01}%
\psfrag{v03}[r][r][0.75]{0.02}%
\psfrag{v04}[r][r][0.75]{0.03}%
\psfrag{v05}[r][r][0.75]{0.04}%
\psfrag{v06}[r][r][0.75]{0.05}%
\psfrag{v07}[r][r][0.75]{0.06}%
\psfrag{v08}[r][r][0.75]{0.07}%
\psfrag{v09}[r][r][0.75]{0.08}%
\psfrag{v10}[r][r][0.75]{0.09}%
\psfrag{v11}[r][r][0.75]{0.1}%
\psfrag{v12}[r][r][0.75]{0}%
\psfrag{v13}[r][r][0.75]{0.02}%
\psfrag{v14}[r][r][0.75]{0.04}%
\psfrag{v15}[r][r][0.75]{0.06}%
\psfrag{v16}[r][r][0.75]{0.08}%
\psfrag{v17}[r][r][0.75]{0.1}%
%
\psfrag{z01}[r][r][0.75]{400}%
\psfrag{z02}[r][r][0.75]{600}%
\psfrag{z03}[r][r][0.75]{800}%
\psfrag{z04}[r][r][0.75]{400}%
\psfrag{z05}[r][r][0.75]{500}%
\psfrag{z06}[r][r][0.75]{600}%
\psfrag{z07}[r][r][0.75]{700}%
\psfrag{z08}[r][r][0.75]{800}%
  \includegraphics[width=8cm]{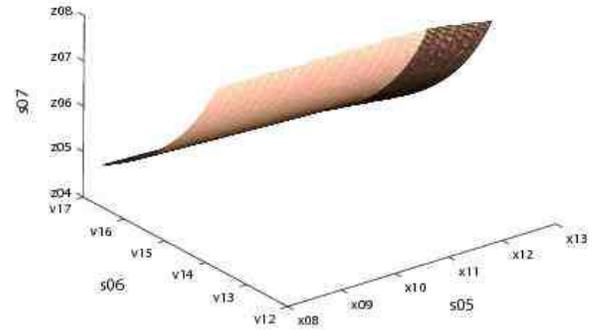}\\
  \caption{Pareto frontier of Slide-o-Cam mechanisms}
  \label{fig:3DParetoFinal}
\end{figure}

The optimization problem defined in Section~\ref{sec:pbformulation}
is multiobjective with objective functions of a different nature.
For this reason, the optimum solutions of the problem can be
illustrated by means of a Pareto frontier \cite{Collette:2006}. As
the problem involves three objective functions, i.e., $\mu_{max}$,
$P_{max}$ and $S_M$, the corresponding Pareto frontier is depicted
in 3D space as shown in Figs.~\ref{fig:3DPareto23cams} and
\ref{fig:3DParetoFinal}. Figure~\ref{fig:3DPareto23cams} illustrates
the Pareto frontiers of a two- and a three-conjugate cam mechanisms.
As we want to minimize the three objective functions concurrently,
the closer the Pareto frontier to the origin, the better the design.
In Fig.~\ref{fig:3DPareto23cams}, we notice that the optimum
solutions obtained with a three-conjugate cam mechanism are slightly
better when $\mu_{max}$ is smaller than $24^\circ$. Otherwise, a
two-conjugate cam mechanism turns out to be more interesting.
Nevertheless, the difference between the optimum solutions obtained
with a two- and a three-conjugate cam mechanisms remains low.
Figure~\ref{fig:3DParetoFinal} depicts the region closest to the
origin of the two frontiers shown in Fig.~\ref{fig:3DPareto23cams}.
It also shows the Pareto frontier of Slide-o-Cam mechanisms,
regardless of the number of conjugate-cams.
\begin{figure}[!h]
\centering
\psfrag{s02}[t][t]{$\mu_{max}$~[deg]}
\psfrag{s03}[b][b]{$S_M$~[m]}%
\psfrag{s04}[r][r]{$P_{max}$~[MPa]}%
%
\psfrag{x01}[t][t][0.75]{5}%
\psfrag{x02}[t][t][0.75]{10}%
\psfrag{x03}[t][t][0.75]{15}%
\psfrag{x04}[t][t][0.75]{20}%
\psfrag{x05}[t][t][0.75]{25}%
\psfrag{x06}[t][t][0.75]{30}%
%
\psfrag{v01}[r][r][0.75]{0}%
\psfrag{v02}[r][r][0.75]{0.01}%
\psfrag{v03}[r][r][0.75]{0.02}%
\psfrag{v04}[r][r][0.75]{0.03}%
\psfrag{v05}[r][r][0.75]{0.04}%
\psfrag{v06}[r][r][0.75]{0.05}%
\psfrag{v07}[r][r][0.75]{0.06}%
\psfrag{v08}[r][r][0.75]{0.07}%
\psfrag{v09}[r][r][0.75]{0.08}%
\psfrag{v10}[r][r][0.75]{0.09}%
  \includegraphics[width=8cm]{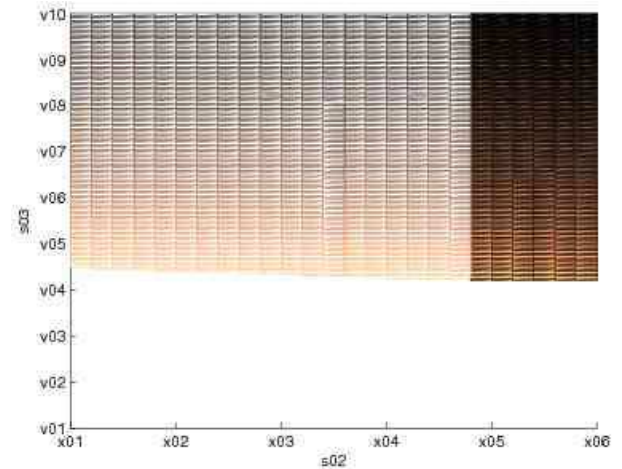}\\
  \caption{Pareto frontier w.r.t. $\mu_{max}$ and $S_M$}
  \label{fig:ParetomuSM}
\end{figure}

\begin{figure}[!h]
\centering
\psfrag{s02}[t][t]{$\mu_{max}$~[deg]}
\psfrag{s03}[b][b]{$S_M$~[m]}%
\psfrag{s04}[r][r]{$P_{max}$~[MPa]}%
%
\psfrag{x01}[t][t][0.75]{5}%
\psfrag{x02}[t][t][0.75]{10}%
\psfrag{x03}[t][t][0.75]{15}%
\psfrag{x04}[t][t][0.75]{20}%
\psfrag{x05}[t][t][0.75]{25}%
\psfrag{x06}[t][t][0.75]{30}%
%
\psfrag{z01}[r][r][0.75]{450}%
\psfrag{z02}[r][r][0.75]{500}%
\psfrag{z03}[r][r][0.75]{550}%
\psfrag{z04}[r][r][0.75]{600}%
\psfrag{z05}[r][r][0.75]{650}%
\psfrag{z06}[r][r][0.75]{700}%
\psfrag{z07}[r][r][0.75]{750}%
\psfrag{z08}[r][r][0.75]{800}%
  \includegraphics[width=8cm]{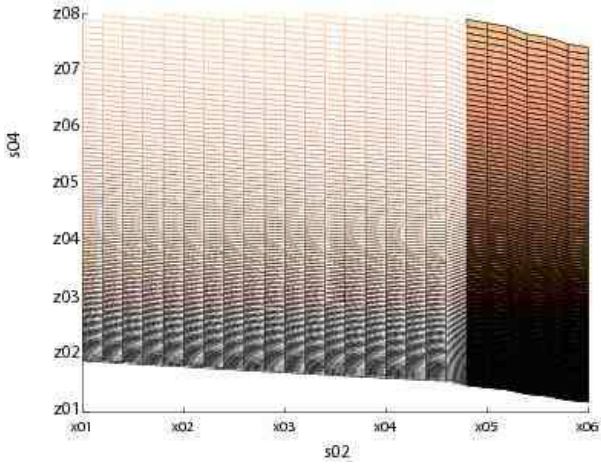}\\
  \caption{Pareto frontier w.r.t. $P_{max}$ and $\mu_{max}$}
  \label{fig:ParetoPhmu}
\end{figure}

\begin{figure}[!h]
\centering
\psfrag{s02}[t][t]{$\mu_{max}$~[deg]}
\psfrag{s03}[b][b]{$S_M$~[m]}%
\psfrag{s04}[c][c]{$P_{max}$~[MPa]}%
\psfrag{v01}[r][r][0.75]{0}%
\psfrag{v02}[r][r][0.75]{0.01}%
\psfrag{v03}[r][r][0.75]{0.02}%
\psfrag{v04}[r][r][0.75]{0.03}%
\psfrag{v05}[r][r][0.75]{0.04}%
\psfrag{v06}[r][r][0.75]{0.05}%
\psfrag{v07}[r][r][0.75]{0.06}%
\psfrag{v08}[r][r][0.75]{0.07}%
\psfrag{v09}[r][r][0.75]{0.08}%
\psfrag{v10}[r][r][0.75]{0.09}%
%
\psfrag{z01}[r][r][0.75]{450}%
\psfrag{z02}[r][r][0.75]{500}%
\psfrag{z03}[r][r][0.75]{550}%
\psfrag{z04}[r][r][0.75]{600}%
\psfrag{z05}[r][r][0.75]{650}%
\psfrag{z06}[r][r][0.75]{700}%
\psfrag{z07}[r][r][0.75]{750}%
\psfrag{z08}[r][r][0.75]{800}%
  \includegraphics[width=8cm]{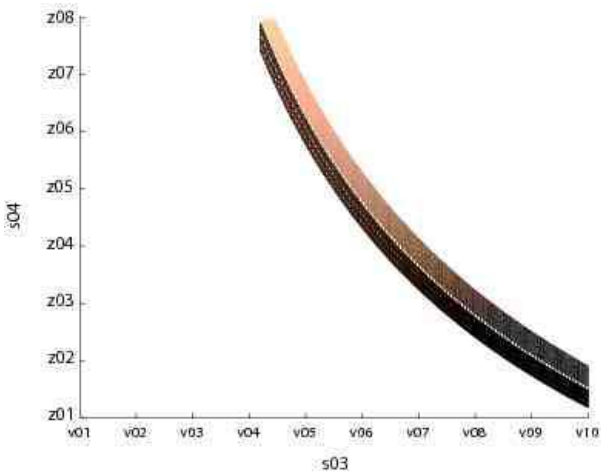}\\
  \caption{Pareto frontier w.r.t. $P_{max}$ and $S_M$}
  \label{fig:ParetoPhSM}
\end{figure}
For better clarity of the results, Figs.~\ref{fig:ParetomuSM},
\ref{fig:ParetoPhmu} and \ref{fig:ParetoPhSM} illustrate the
projections of the Pareto frontier shown in
Fig.~\ref{fig:3DParetoFinal} w.r.t $\mu_{max}$ and $S_M$; $P_{max}$
and $\mu_{max}$; and $P_{max}$ and $S_M$, respectively. These
figures allow us to see clearly the location the optimum and the
feasible solutions of the problem at hand.
\begin{figure}[!h]
\centering
\psfrag{phics}[t][t]{$d_{cs}$~[mm]}
\psfrag{toto}[b][b]{$r$~[mm]}%
\psfrag{M1}[c][c]{$M_1$}%
\psfrag{M2}[c][c]{$M_2$}%
\psfrag{opt}[c][c]{Optimal solutions}
  \includegraphics[width=8cm]{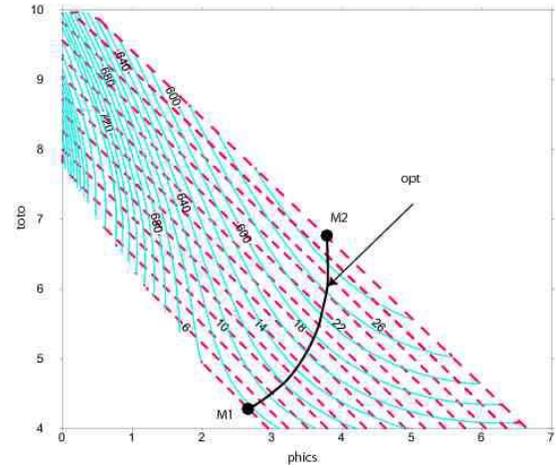}\\
  \caption{Contours of $\mu$ and $P$ w.r.t $d_{cs}$ and $r$ and the location of the optimal solutions for a two conjugate-cam mechanism with $S_M=0.06$~m}
  \label{fig:2camcontour}
\end{figure}

\begin{figure}[!h]
\centering
\psfrag{phics}[t][t]{$d_{cs}$~[mm]}
\psfrag{toto}[b][b]{$r$~[mm]}%
\psfrag{opt}[c][c]{Optimal solutions}
\psfrag{M1}[c][c]{$M_3$}%
\psfrag{M2}[c][c]{$M_4$}%
  \includegraphics[width=8cm]{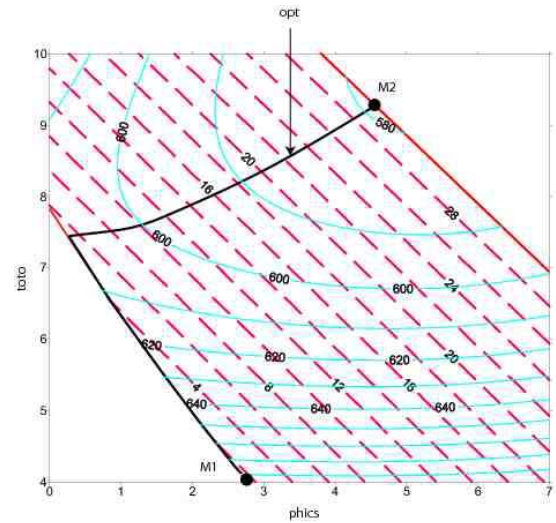}\\
  \caption{Contours of $\mu$ and $P$ w.r.t. $d_{cs}$ and $r$ and the location of the optimal solutions for a three conjugate-cam mechanism with $S_M=0.06$~m}
  \label{fig:3camcontour}
\end{figure}
Figures~\ref{fig:2camcontour} and~\ref{fig:3camcontour} illustrate
the contours of $\mu_{max}$ and $P_{max}$ w.r.t $d_{cs}$ and $r$ for
a two- and a three-conjugate cam mechanisms with $S_M=0.06$~m. On
the one hand, the continuous lines depict the iso-contours of
$\mu_{max}$. On the other hand, the broken lines depict the
iso-contours of $P_{max}$. Besides, Figs.~\ref{fig:2camcontour}
and~\ref{fig:3camcontour} highlight the location of the optimum
solutions for a two- and a three-conjugate cam mechanisms with
$S_M=0.06$~m. We can notice that the line of optimum solutions in
the space of design variables $d_{cs}$ and $r$ is longer in
Fig.~\ref{fig:3camcontour} than in Fig.~\ref{fig:2camcontour}. This
means that a three-conjugate cam mechanism allows more optimal
solutions than its two-conjugate cam counterpart. In this vein, it
is more interesting to design a three-conjugate cam mechanism.

\begin{figure}[!h]
\centering
\psfrag{M1}[c][c]{$M_1$}%
\psfrag{M2}[c][c]{$M_2$}%
\psfrag{ph1}[c][c][0.6]{$P_{max}=653.83$~MPa}%
\psfrag{ph2}[c][c][0.6]{$P_{max}=562.12$~MPa}%
\psfrag{mu1}[c][c][0.6]{$\mu_{max}=3^{\circ}$}%
\psfrag{mu2}[c][c][0.6]{$\mu_{max}=30^{\circ}$}%
  \includegraphics[width=8cm]{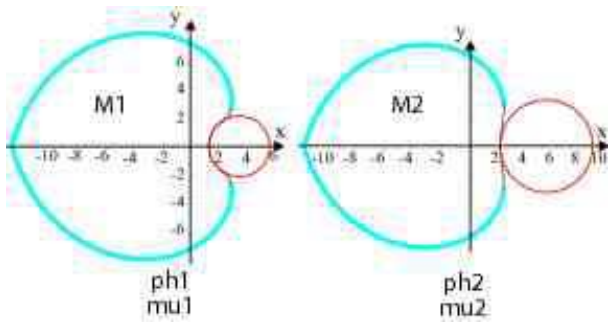}\\
  \caption{Optimal two conjugate-cam mechanisms}
  \label{fig:2camexample}
\end{figure}

Figure~\ref{fig:2camexample} depicts the mechanisms corresponding to
points $M_1$ and $M_2$ that are plotted in
Fig.\ref{fig:2camcontour}. For $M_1$, $d_{cs}=2.6$~mm, $r=4.24$~mm,
$\mu_{max}=3^{\circ}$ and $P_{max}=653.83$~MPa. For $M_2$,
$d_{cs}=4.16$~mm, $r=6.4$~mm, $\mu_{max}=30^{\circ}$ and
$P_{max}=562.12$~MPa.

\begin{figure}[!h]
\centering
\psfrag{M1}[c][c]{$M_3$}%
\psfrag{M2}[c][c]{$M_4$}%
\psfrag{ph1}[c][c][0.6]{$P_{max}=654.57$~MPa}%
\psfrag{ph2}[c][c][0.6]{$P_{max}=579.45$~MPa}%
\psfrag{mu1}[c][c][0.6]{$\mu_{max}=2^{\circ}$}%
\psfrag{mu2}[c][c][0.6]{$\mu_{max}=30^{\circ}$}%
  \includegraphics[width=8cm]{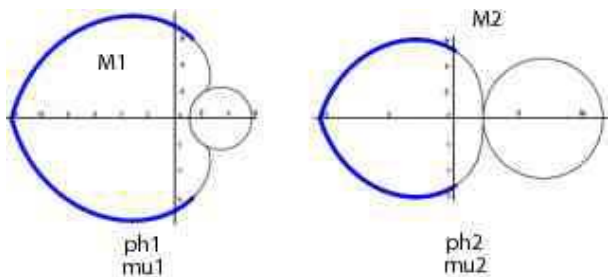}\\
  \caption{Optimal three conjugate-cam mechanisms}
  \label{fig:3camexample}
\end{figure}

Figure~\ref{fig:3camexample} depicts the mechanisms corresponding to
points $M_3$ and $M_4$ that are plotted in
Fig.\ref{fig:3camcontour}. For $M_3$, $d_{cs}=2.2$~mm, $r=4.68$~mm,
$\mu_{max}=2^{\circ}$ and $P_{max}=654.57$~MPa. For $M_4$,
$d_{cs}=4.56$~mm, $r=9.28$~mm, $\mu_{max}=30^{\circ}$ and
$P_{max}=579.45$~MPa.

According to Figs.~\ref{fig:2camcontour} and \ref{fig:3camcontour},
we can notice that the higher $r$, the smaller $P_{max}$. Indeed,
the maximum Hertz pressure values corresponding to $M_2$ and $M_4$
are smaller than the ones corresponding to $M_1$ and $M_3$. However,
the size of the mechanism along the $x$-axis is higher for $M_2$ and
$M_4$. Moreover, this induces a better transmission of the torque as
$d_{cs}$ is higher. Finally, we can notice that the profiles of
$M_2$ and $M_4$ are easier to machine as they are fully convex.

\section{CONCLUSIONS}
The multiobjective optimization of Slide-o-Cam was reported in this
paper. Slide-o-Cam is a cam mechanism with multiple rollers mounted
on a common translating follower. This transmission provides
pure-rolling motion, thereby reducing the friction of
rack-and-pinions and linear drives. A Pareto frontier was obtained
by means of a multiobjective optimization. This optimization is
based on three objective functions: (i)~the pressure angle, which is
a suitable performance index for the transmission because it
determines the amount of force transmitted to the load vs.\ that
transmitted to the machine frame; (ii)~the Hertz pressure used to
evaluate the stresses produced in the contact surface between the
cams and the rollers; and (iii)~the size of the mechanism
characterized by the number of cams and their width. It turns out
that three-conjugate cam mechanisms have globally better performance
that their two-conjugate cam counterparts. However, the difference
is small.


%
%
%
%

\end{document}